\documentclass[letterpaper,10pt,conference]{ieeeconf}

\usepackage{graphics} 
\usepackage{amsmath} 
\usepackage{amssymb}  
\usepackage{cite}
\usepackage{color}
\usepackage{graphicx}
\usepackage{subcaption}
\usepackage{booktabs}
\usepackage{subfig}

\title{\LARGE \bf
Evict3R: Training-Free Token Eviction for Memory-Bounded Streaming Visual Geometry  Transformers
}

\author{
Soroush Mahdi\textsuperscript{1},
Fardin Ayar\textsuperscript{1},
Ehsan Javanmardi\textsuperscript{2},
Manabu Tsukada\textsuperscript{2}
Mahdi Javanmardi\textsuperscript{1}\textsuperscript{*},
\\[0.5em]
\textsuperscript{1}Amirkabir University of Technology (AUT), Tehran, Iran\\
\textsuperscript{2}The University of Tokyo, Tokyo, Japan\\[0.5em]
\textsuperscript{*}\textit{Corresponding author: mjavan@aut.ac.ir}
}

\begin{document}

\maketitle
\thispagestyle{empty}
\pagestyle{empty}

\begin{abstract}
Streaming visual transformers like StreamVGGT achieve strong 3D perception but suffer from unbounded growth of key–value (KV) memory, which limits scalability. We propose a training-free, inference-time token eviction policy that bounds memory by discarding redundant tokens while keeping the most informative ones. Our method uses significantly less memory with little to no drop in accuracy: on 7-Scenes with long sequences it reduces peak memory from 18.63\,GB to 9.39\,GB while accuracy and completeness drop by only 0.003. Under strict memory budgets, eviction enables denser frame sampling, which improves reconstruction accuracy compared to the baseline. Experiments across video depth estimation (Sintel, KITTI), 3D reconstruction (7-Scenes, NRGBD), and camera pose estimation (Sintel, TUM-dynamics) show that our approach closely matches StreamVGGT at a fraction of the memory and makes long-horizon streaming inference more practical.
\end{abstract}

\section{Introduction}

Robotic platforms rely on dense, view-consistent 3D perception for navigation, manipulation, and long-horizon autonomy. Classical Structure-from-Motion (SfM) and Multi-View Stereo (MVS) pipelines deliver accurate reconstructions via global optimization \cite{hartley2000multiple, furukawa2015multi}, but their iterative bundle adjustment and repeated data association hinder real-time operation and incremental updates on resource-constrained hardware. Recent learning-based approaches replace hand-crafted geometry with end-to-end predictors that infer depth, pose, and point maps directly from images \cite{wang2024dust3r, Yang2025Fast3RT3, Wang2025VGGTVG}. Among these, large feed-forward transformers achieve strong accuracy by reasoning jointly across many views, but they are typically offline: each new frame requires re-encoding the entire sequence, making them ill-suited to streaming robotics workloads.

Causal, streaming variants have emerged to close this gap. StreamVGGT adopts the autoregressive paradigm of language models: each incoming frame attends to a cached key–value (KV) memory of all prior frames, avoiding redundant recomputation and enabling low-latency updates \cite{Zhuo2025Streaming4V}. However, the same KV cache that boosts throughput also causes a fundamental scalability bottleneck: its size increases with sequence length. Even with fewer than 10 frames, the KV cache size can exceed the model size! Other streaming designs impose fixed external memories or explicit spatial pointer stores \cite{Wang_2025_CVPR, wang20243d, Wu2025Point3RS3}, but capacity limits introduce information loss and drift, and the required architectural changes complicate adoption.

 We ask: \emph{Can a streaming visual transformer preserve the benefits of long-range context while bounding its internal KV memory, without retraining or modifying model weights?}

\subsubsection{Approach} We introduce a training-free, inference-time token budgeting and eviction mechanism for StreamVGGT that (i) enforces \emph{per-layer} KV-cache budgets and (ii) selects tokens for retention using an attention-based importance score. We further allocate the total budget across global-attention layers using an attention-sparsity prior, reflecting the empirical observation that early/late layers attend more densely than middle layers, in line with long-context LLM findings \cite{Zhang2023H2OHO, Liu2023ScissorhandsET, Wan2024D2ODD, cai2024pyramidkv}. 

\begin{figure*}[h]
  \centering
  \begin{subfigure}{0.48\linewidth}
    \includegraphics[width=\linewidth]{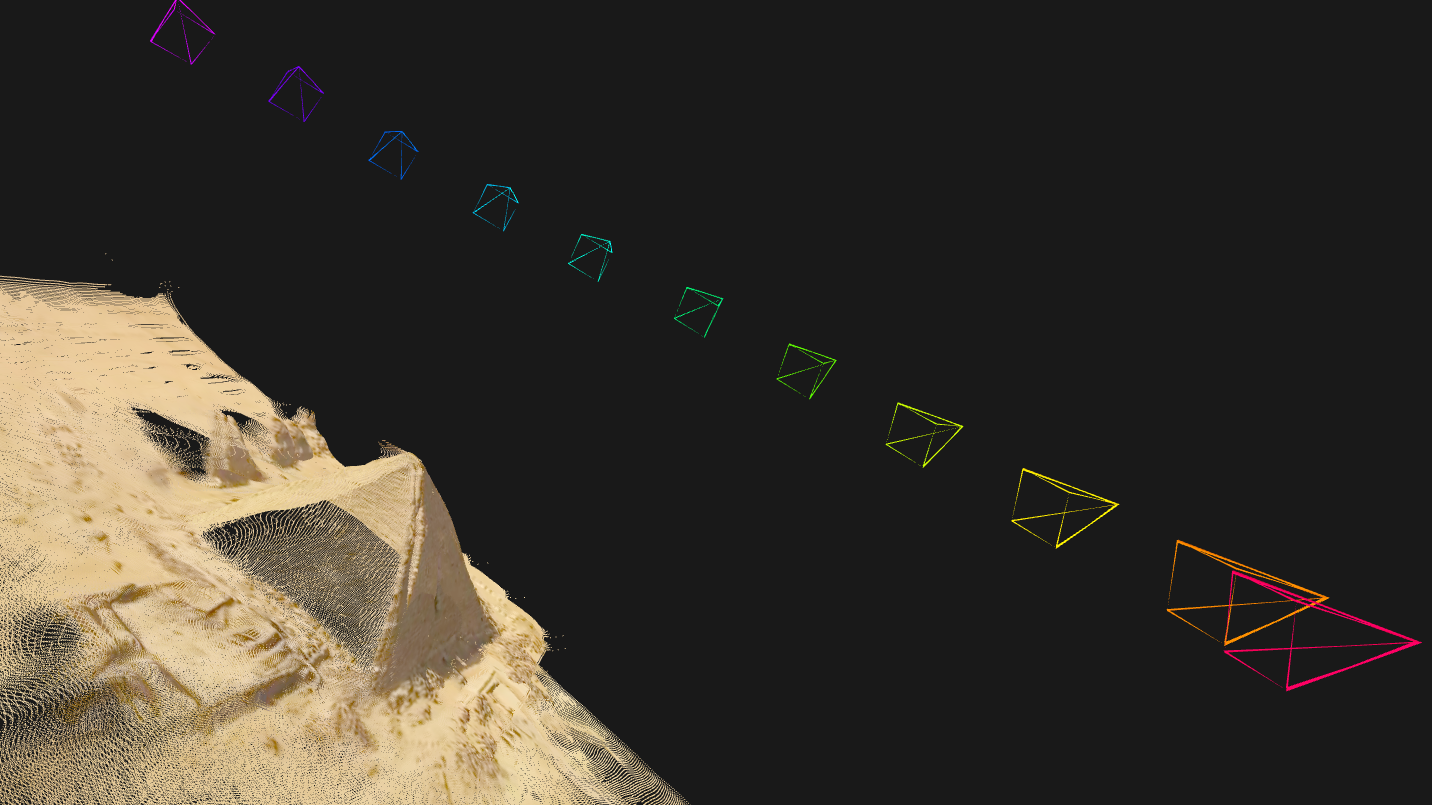}
    \caption{StreamVGGT}
  \end{subfigure}\hfill
  \begin{subfigure}{0.48\linewidth}
    \includegraphics[width=\linewidth]{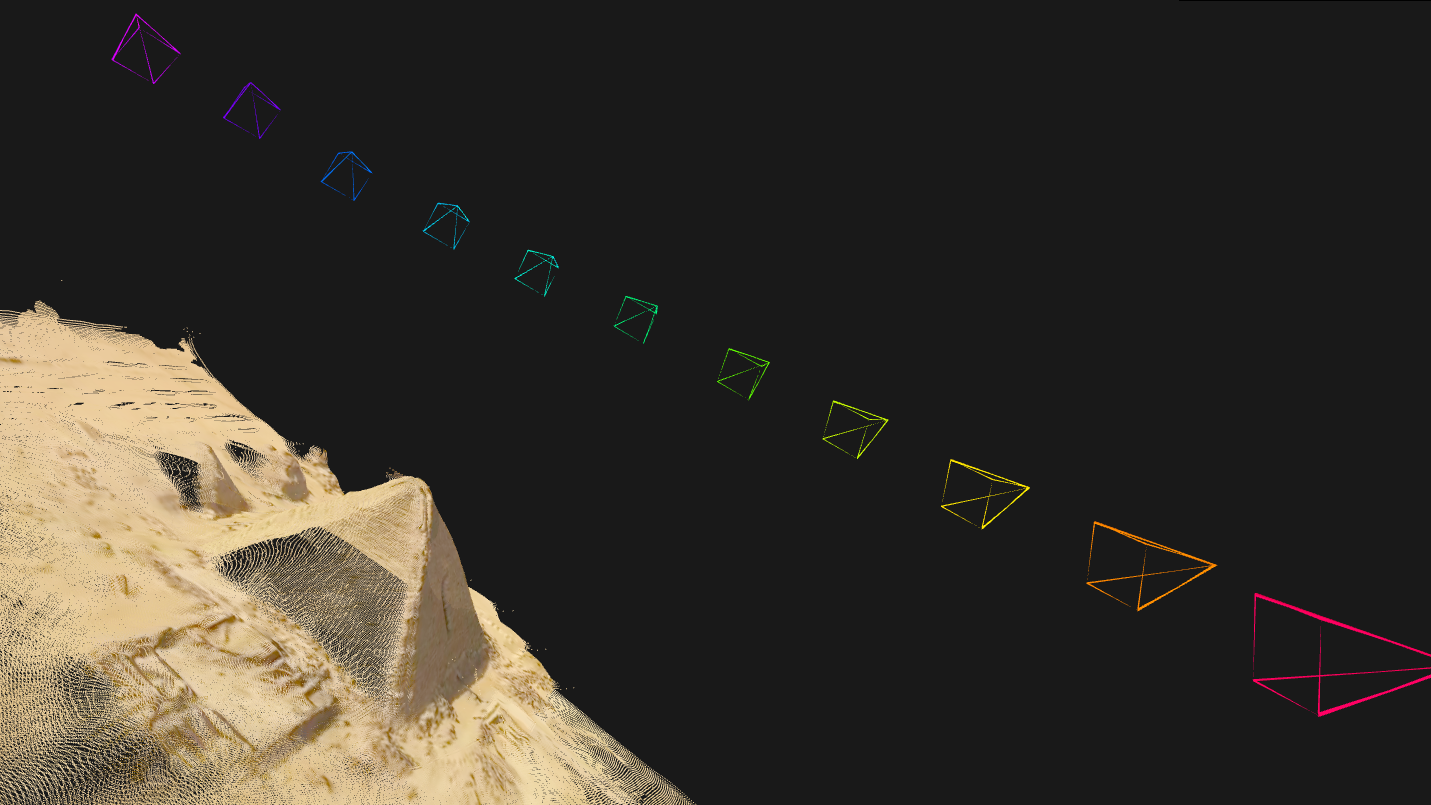}
    \caption{Ours ($B{=}0.2$)}
  \end{subfigure}

  \caption{\textbf{Qualitative 3D pointmap comparison.} Reconstructions from StreamVGGT and our method with $B{=}0.2$. Despite using less than half the memory, our method shows no visible drop in quality.}

  \label{fig:pointmap_comparison}
\end{figure*}

Our experiments show that on video depth, multi-view reconstruction, and camera pose, our method matches StreamVGGT accuracy at moderate budgets while substantially reducing peak memory and per-frame latency; critically, it enables much longer sequences (e.g., $2{\times}$–$10{\times}$) on a single GPU. Also, qualitative reconstructions remain complete and consistent under tight budgets (Fig.~\ref{fig:pointmap_comparison}).

\section{Related Work}
 
\subsubsection{From Conventional to Learning-Based Reconstruction}
Classical 3D reconstruction pipelines, including Structure-from-Motion (SfM) and Multi-View Stereo (MVS), achieve high accuracy through sophisticated global optimization procedures \cite{hartley2000multiple, furukawa2015multi}.
These methods, however, suffer from computational limitations that restrict their applicability to scenarios requiring real-time processing or incremental updates.
In dynamic environments or scenes with limited texture, feature matching and photometric consistency deteriorate, further compromising system robustness.
These fundamental limitations have motivated the development of end-to-end learning-based approaches that directly infer 3D geometry from visual inputs.
Pairwise reconstruction methods, exemplified by DUSt3R \cite{wang2024dust3r} demonstrate the capability to predict view-consistent point maps from image pairs.
While effective for two-view scenarios, pairwise approaches inherently lack the global context necessary for handling complex multi-view sequences with occlusions and varying viewpoints.
Large-scale feed-forward transformers, such as Fast3R \cite{Yang2025Fast3RT3} and VGGT \cite{Wang2025VGGTVG}, address this limitation by enabling comprehensive multi-view reasoning through single forward passes that consider all views simultaneously.
However, these transformer-based models operate exclusively in offline mode, requiring complete sequence re-encoding whenever new frames arrive, which fundamentally prevents real-time streaming applications.

\subsubsection{Streaming 4D Reconstruction}
Real-time 4D reconstruction has motivated streaming methods that process continuous video with a persistent state while remaining efficient.
Memory-augmented approaches such as \emph{CUT3R} \cite{Wang_2025_CVPR} maintain \emph{explicit} spatial state to predict pointmaps in a common frame.
More recent works—\emph{Spann3R} \cite{wang20243d} and \emph{Point3R} \cite{Wu2025Point3RS3}—introduce stronger memory management at the cost of architectural complexity.
\emph{Spann3R} employs a two-tier memory (dense working + sparse long-term) and \emph{evites} its long-term store by retaining top-$k$ items according to \emph{accumulated attention}, keeping the long-term memory bounded.
\emph{Point3R} maintains an explicit \emph{spatial pointer memory} and fuses nearby pointers under a distance schedule to suppress redundancy as the scene grows; this curbs growth but does not guarantee a fixed bound.
\emph{StreamVGGT} \cite{Zhuo2025Streaming4V} takes another path: temporal \emph{causal attention} with a cached KV memory enables efficient autoregressive updates, but the cache grows \emph{linearly} with sequence.

We manage \emph{KV memory} of StreamVGGT with a \emph{training-free, inference-time} policy that enforces \emph{per-layer token budgets}, ranks keys by attention with \emph{exposure/length} normalization, and preserves a fixed set of important tokens.

Compared to \emph{ v}’s attention-based eviction, ours adds \emph{per-layer} control and the mentioned normalizations, remaining plug-and-play.
In contrast to \emph{Point3R}'s explicit pointer memory, our method is architecture-agnostic and enforces a strict, per-layer memory budget. By operating directly on the key and value (KV) cache within StreamVGGT, our approach avoids the need to reproject features into keys and values at each step, resulting in lower computational overhead and seamless integration with existing transformer-based architectures.

\begin{figure*}[t]
  \centering
    \includegraphics[width=0.9\textwidth]{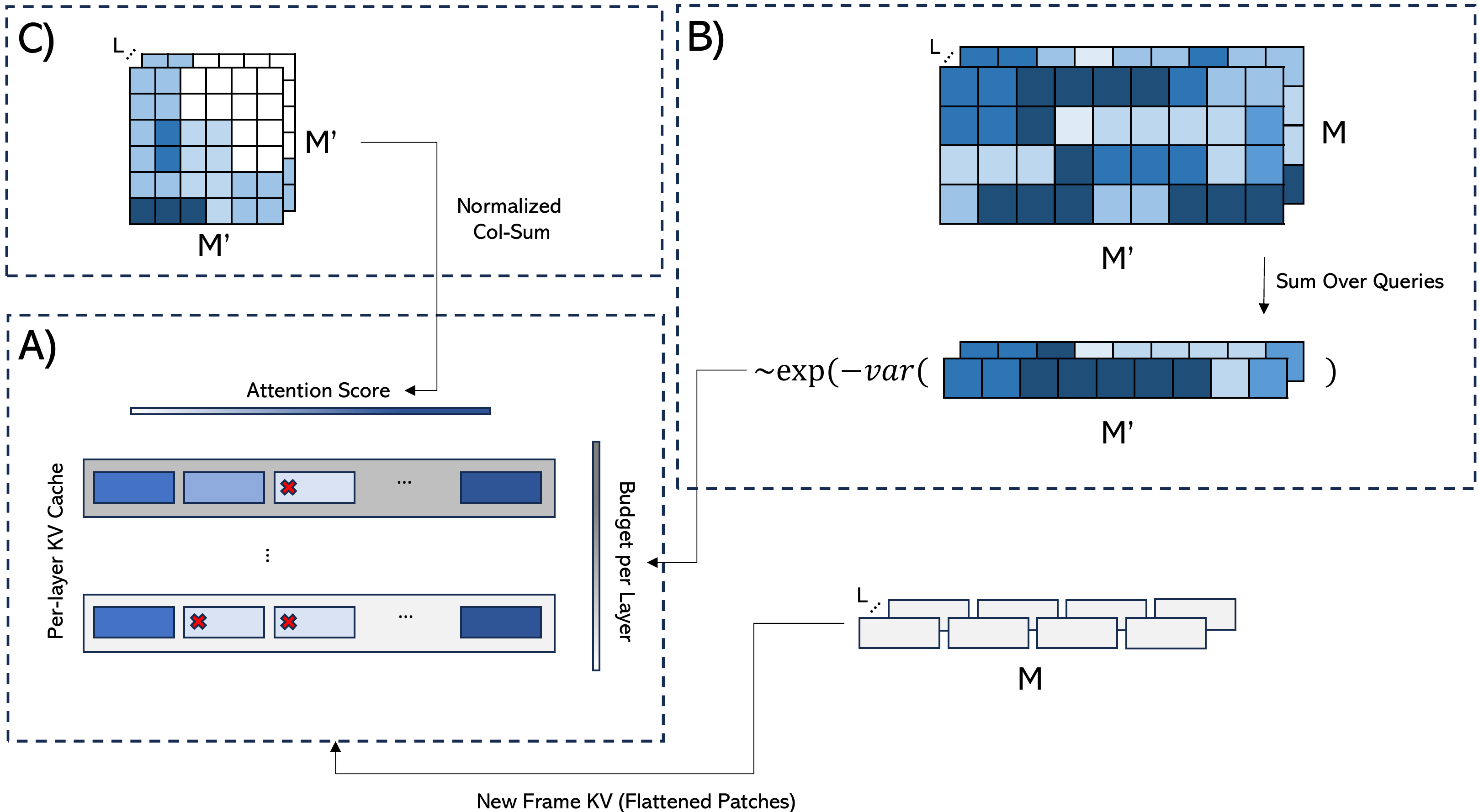}

  \caption{\textbf{Overview of our token eviction framework.} 
(A) Tokens in each layer’s KV cache are ranked by attention importance and evicted under a per-layer budget. 
(B) Importance scores are derived from query–key attention statistics. 
(C) Scores are normalized to guide consistent token selection across layers.}

  \label{fig:method_overview}
\end{figure*}

\subsubsection{Token Management and Eviction}
Token reduction is essential for managing the computational demands of transformer architectures. In computer vision, early methods such as token merging (e.g., ToMe \cite{Bolya2022TokenMY}) and token pruning (e.g., DynamicViT \cite{Rao2021DynamicViTEV}) were developed to reduce redundancy and focus computation on informative tokens. More recent approaches combine these strategies, as in PuMer \cite{Cao2023PuMerPA}, to further improve efficiency.

Key-value (KV) caching is a core technique in transformer models with causal attention, such as those used in large language models (LLMs). In causal attention, each new token attends only to previous tokens, enabling efficient autoregressive inference by storing and reusing the keys and values (the "KV cache") from earlier steps. This approach allows models to process long sequences incrementally without recomputing attention over the entire history at each step. However, as the sequence grows, the KV cache can become prohibitively large, leading to unbounded memory usage. To address this, the LLM community has developed a range of token eviction and memory budgeting strategies for managing KV cache growth during long-context inference. Early methods such as H2O \cite{Zhang2023H2OHO} and Scissorhands \cite{Liu2023ScissorhandsET} use attention-based importance scores to identify and retain the top-$k$ most relevant tokens, evicting less important ones. More recent approaches, including D2O \cite{Wan2024D2ODD} and PyramidKV \cite{cai2024pyramidkv}, introduce layer-wise KV cache budget allocation, recognizing that different transformer layers exhibit varying degrees of attention sparsity and thus require different memory budgets.

Since our baseline method (StreamVGGT \cite{Zhuo2025Streaming4V}) adopts the LLM paradigm with causal attention and cached tokens, leveraging these advanced eviction strategies is particularly relevant. To our knowledge, no prior work has successfully integrated principled token eviction from the LLM literature into causal visual transformers for scalable and accurate streaming reconstruction.

\section{Preliminaries \& Motivation}

This section reviews VGGT's offline multi-view reconstruction capabilities and StreamVGGT's causal streaming adaptation.
We examine how StreamVGGT enables real-time processing through key-value caching while highlighting its unbounded memory growth limitations.
This analysis motivates our solution for addressing the scalability challenges that restrict the practical deployment of streaming reconstruction systems.

\subsection{VGGT \& StreamVGGT: Offline vs Online Multi-View Reconstruction}

\textbf{VGGT} processes $N$ input images $\{I_t\}_{t=1}^{T}$,  $I_t \in \mathbb{R}^{H \times W \times 3}$ through DINO encoder \cite{oquab2023dinov2} to produce feature tokens $F_t = \mathrm{Enc}(I_t) \in \mathbb{R}^{M \times d}$ where $M$ is the number of tokens per frame and $d$ is the token dimension.
These tokens are concatenated into $Z^{(0)} = \mathrm{Concat}(F_1,\ldots,F_N) \in \mathbb{R}^{NM \times d}$ and processed by a transformer with alternating attention layers.
Each layer applies frame-wise self-attention $\mathcal{F}$ followed by global self-attention $\mathcal{G}$: $Z^{(\ell+1)} = \mathcal{G}(\mathcal{F}(Z^{(\ell)}))$ for $\ell = 0,\ldots,L-1$ where $L$ is the total number of layers.
The final features are decoded by per-view DPT head \cite{ranftl2021vision} to produce geometry, depth, and pose estimates: $(g_t, D_t, P_t, T_t) = \mathrm{DPT}(Z_t^{(L)})$ where $g_t \in \mathbb{R}^{9}$ represents camera parameters (intrinsics and extrinsics), $D_t \in \mathbb{R}^{H \times W}$ denotes the predicted depth map for view $t$, $P_t \in \mathbb{R}^{3 \times H \times W}$ is the point map providing dense 3D coordinates for each pixel, and $T_t \in \mathbb{R}^{C \times H \times W}$ represents a grid of $C$-dimensional features for point tracking. While effective, VGGT's global attention couples all views, resulting in quadratic computational complexity $O(N^2M^2)$ and requiring complete re-encoding when new frames arrive, making it infeasible for robotic applications requiring real-time processing and incremental updates.

\begin{figure*}[t]
  \centering
  \begin{subfigure}{0.24\textwidth}
    \includegraphics[width=\linewidth]{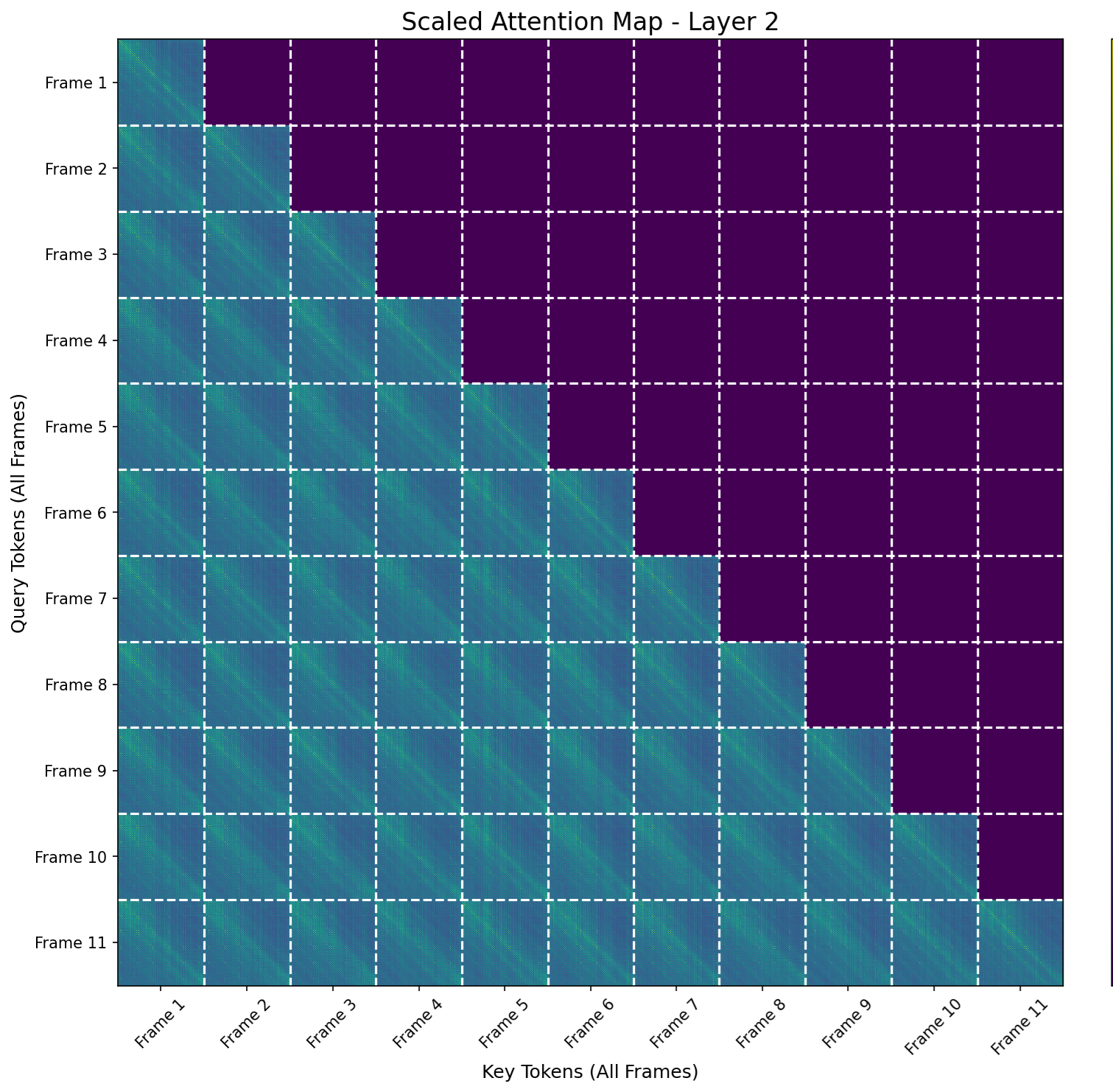}
    \caption{Layer 2}
  \end{subfigure}\hfill
  \begin{subfigure}{0.24\textwidth}
    \includegraphics[width=\linewidth]{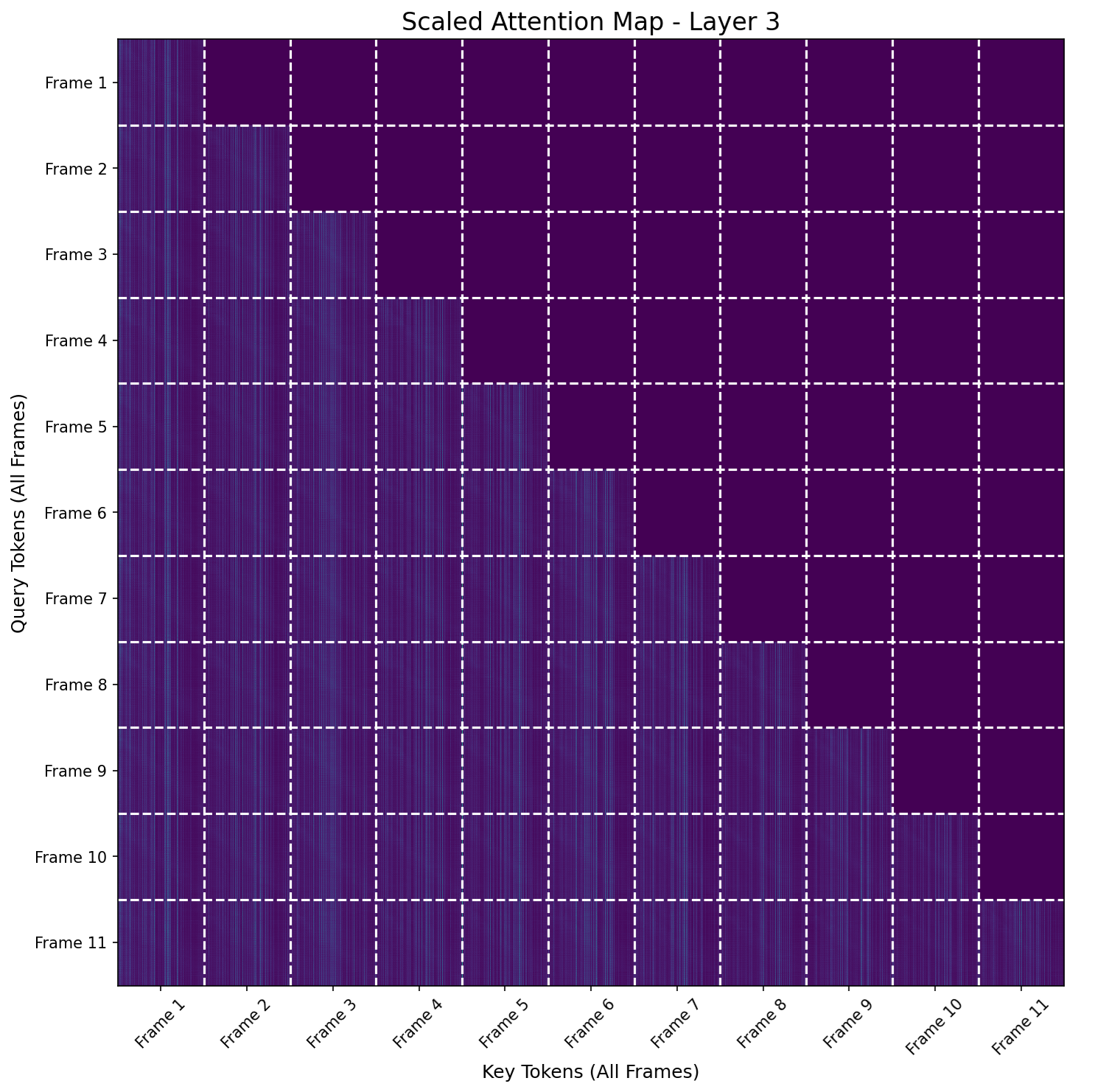}
    \caption{Layer 3}
  \end{subfigure}\hfill
  \begin{subfigure}{0.24\textwidth}
    \includegraphics[width=\linewidth]{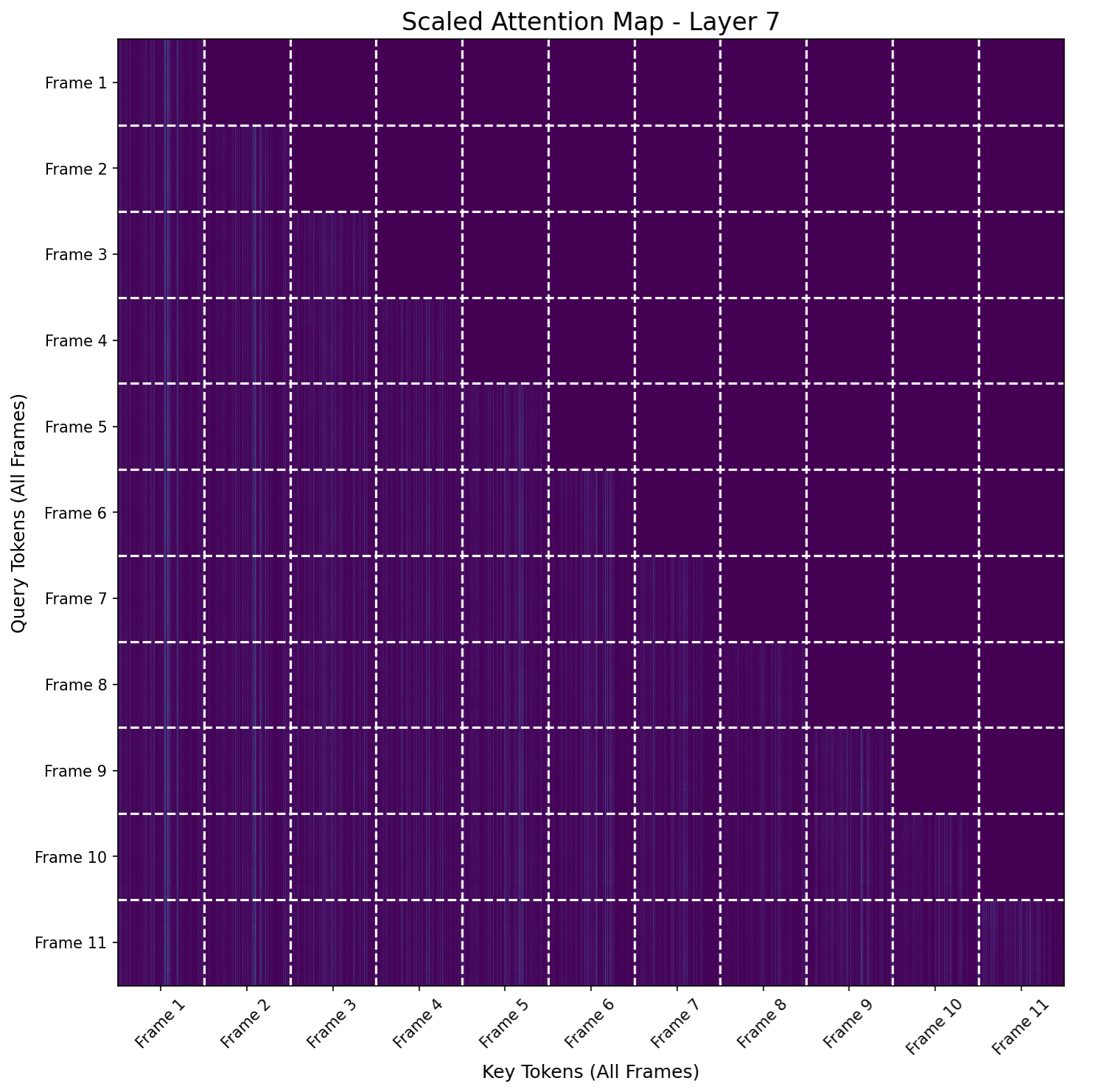}
    \caption{Layer 7}
  \end{subfigure}\hfill
  \begin{subfigure}{0.24\textwidth}
    \includegraphics[width=\linewidth]{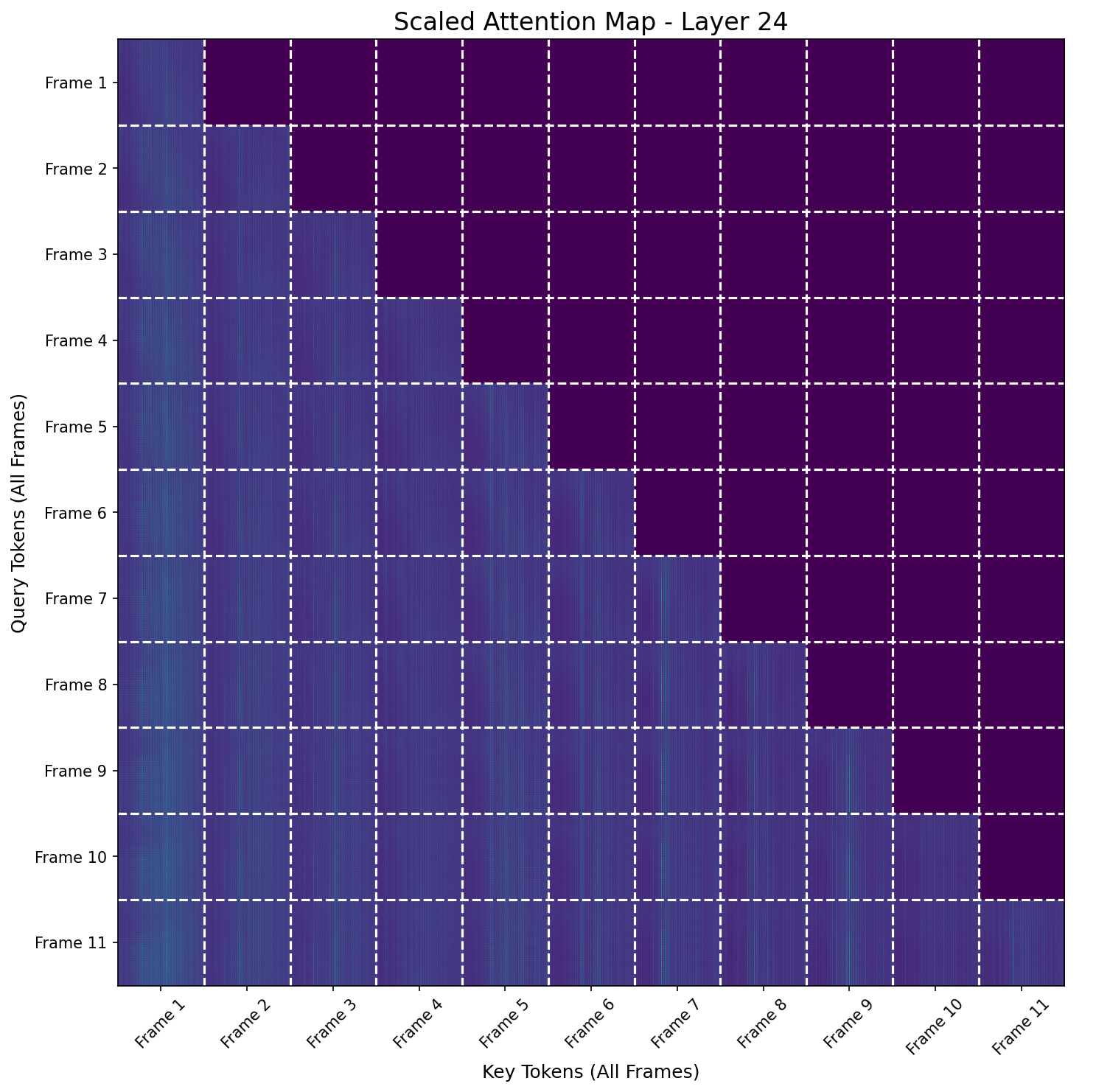}
    \caption{Layer 24}
  \end{subfigure}



  \caption{%
    \textbf{Head-averaged attention maps across temporal layers in StreamVGGT\cite{Zhuo2025Streaming4V}.}
    Each subplot corresponds to one layer, averaged across all attention heads.
    Color scales are independent per subplot (not comparable across layers).
    For better visibility, rows are multiplied by their frame index $r$,
    since later frames have more tokens and lower raw attention values.
    White dashed lines mark frame boundaries along both axes.
  }
  \label{fig:attn_maps_grid}
\end{figure*}
\textbf{StreamVGGT} extends VGGT for real-time processing by replacing global attention with causal attention and introducing key-value (KV) caching.
For each incoming frame $I_t$, it computes $F_t = \mathrm{Enc}(I_t)$ and applies frame-wise attention $\widetilde{Z}_t^{(\ell)} = \mathcal{F}(Z_t^{(\ell)})$ just like in VGGT.
The difference is that global attention processes the current frame against all previous frames using cached keys and values: 
\[ 
Z_t^{(\ell+1)} = \mathcal{G}_{\triangleleft}(\widetilde{Z}_t^{(\ell)}; K_{1:t}^{(\ell)}, V_{1:t}^{(\ell)})
\]
where $\mathcal{G}_{\triangleleft}$ denotes causal global attention, $K_{1:t}^{(\ell)} \in \mathbb{R}^{tM \times d}$ and $V_{1:t}^{(\ell)} \in \mathbb{R}^{tM \times d}$ are the keys and values respectively.It caches $K_{1:t-1}^{(\ell)}$ and $V_{1:t-1}^{(\ell)}$ from previous steps so that each new frame can efficiently attend to all past context without recomputing attention for earlier frames; at each step, only one frame is processed through the model.

The KV cache is updated by appending new keys and values: $K_{1:t}^{(\ell)} \leftarrow \mathrm{Append}(K_{1:t-1}^{(\ell)}, K_t^{(\ell)})$, and $V_{1:t}^{(\ell)} \leftarrow \mathrm{Append}(V_{1:t-1}^{(\ell)}, V_t^{(\ell)})$.
This approach enables incremental processing with low per-step latency, similar to autoregressive language models, but introduces unbounded memory growth as the cache accumulates tokens from all previous frames.

\subsection{Motivation: The Scalability Challenge}

While KV caching enables efficient streaming by avoiding redundant recomputation, it introduces fundamental scalability limitations.
The cache stores keys and values for every past frame across all global attention layers and for each head, leading to unbounded memory growth.
After processing $T$ frames, the total KV cache memory footprint scales as $T \cdot M \cdot L_g \cdot H \cdot (d + d)$, where $T$ is the sequence length, $M$ is the number of global-attention tokens per frame, $L_g$ is the number of global layers, $H$ is the number of heads in each global attention block, and $d$ is the dimension of the keys and values. This also results in quadratic computational complexity $O(T \cdot M^2 \cdot L_g \cdot H)$ for global attention operations.

Existing approaches to streaming reconstruction face a fundamental trade-off between memory efficiency and context preservation.
Memory-augmented models like CUT3R \cite{Wang_2025_CVPR} maintain fixed-size states but suffer from information loss and drift over long sequences due to their bounded capacity.
StreamVGGT preserves complete context through unbounded caching but becomes computationally intractable as sequence length increases.

\section{Proposed Method}

We introduce a training-free streaming token budgeting and eviction mechanism for StreamVGGT that preserves long-range context under a fixed key–value cache budget: Evict3r.
Our design follows the layer-wise budget allocation driven by attention sparsity \cite{Wan2024D2ODD, cai2024pyramidkv}. 

Let $L_g$ denote the number of global-attention layers. We set a total KV budget $B$ and split it across layers as $\sum_{\ell=1}^{L_g} B_\ell \approx B$. When a new frame with $M$ tokens arrives, StreamVGGT first computes the desired outputs. To update KV cache memory, each layer must admit $M$ new keys and values while respecting $B_\ell$. When a layer is full, we evict the least important cached tokens in each layer until $M$ slots are available in each layer. Note that the budget is only for the KV cache. Following the LLM token eviction literature, we use the cumulative-attention score to rank cached tokens for eviction. The intuition is that the tokens which have recived less attention from other tokens during the generation process are less important and can be evicted.

Now we describe or per-layer budget allocation strategy, followed by the details of the token selection strategy for eviction.

\subsection{Per-Layer Budget Allocation}

As shown in the Fig~\ref{fig:attn_maps_grid}, we observe that the first and last global-attention layers exhibit dense and near-uniform attention, whereas middle layers are sparser and more selective. This is similar to the observation in the token eviction literature \cite{Wan2024D2ODD, cai2024pyramidkv} in LLMs, but contrast with them in the way that they usually observe uniform attention only in the first layers. We also find that a subset of tokens remains consistently important across steps. This is also align with the persistence of importance hypothesis \cite{Liu2023ScissorhandsET}. 

Similar to D2O, we use attention map variance as the notion of sparsity.
Let ${A_t}^{(\ell)} \in R^{M \times M'}$ be the mean attention map of heads of layer $(\ell)$ for the frame $t$. Simply put, it is the attention weight of the last frame tokens with respect to themselves and previous tokens (Excluding evicted tokens: $M'=M\times t - \text{nember of evicted tokens in the layer}$). First, we sum ${A_t}^{(\ell)}$ on the query dimension, which gives us a list with the same length as our cached tokens for each layer $S_t^{(\ell)}$. For each input frame, we define:
\[
\sigma_\ell = -\text{Var}(S_{t}^{(\ell)})
\]
We then form layer importance $\pi_\ell$ via a temperature $\tau > 0$:

\[
\pi_\ell = \frac{\exp\big(\sigma_\ell / \tau\big)}{\sum_{r=1}^{L_g} \exp\big(\sigma_r / \tau\big)}\
\]

Then we allocate the budget $B_\ell$ to the layer $\ell$ as

\[
B_\ell = \Big\lfloor B \cdot \pi_\ell \Big\rfloor
\]

\subsection{Token Selection for Eviction}

When a layer's KV cache reaches its limit, we evict the least important cached tokens in the layer until $M$ slots are available.
We rank each cached token $j$ at layer $\ell$ using attention it receives from queries across time, heads, and queries.
We begin with the raw cumulative attention:
\[
c_j^{(\ell)} = \sum_{t \in \mathcal{T}_j} \sum_{h=1}^{H} \sum_{q=1}^{M} a_{t,h,q \rightarrow j}^{(\ell)}\,,
\]
where $c_j^{(\ell)}$ is the cumulative attention received by token $j$ at layer $\ell$, $\mathcal{T}_j$ is the set of time steps during which token $j$ is present in the cache, $H$ is the number of attention heads, $M$ is the number of global tokens per frame, and $a_{t,h,q \rightarrow j}^{(\ell)}$ denotes the attention weight from the $q$-th query token in head $h$ at time $t$ to token $j$ in layer $\ell$.

In contrast to LLM token eviction literature, we argue that row-length normalization is required because early steps can have fewer keys, which would cause their contributions to dominate.
Let $N_t^{(\ell)}$ be the number of keys present when step $t$ was computed, and form the row-length normalized score:
\[
\hat{c}_j^{(\ell)} = \sum_{t \in \mathcal{T}_j} \frac{1}{N_t^{(\ell)}} \sum_{h=1}^{H} \sum_{q=1}^{M} a_{t,h,q \rightarrow j}^{(\ell)}\,.
\]

Also, exposure normalization is required because tokens that remain longer in the cache would accumulate higher scores purely by tenure.
Let $e_j^{(\ell)}$ be the exposure count, i.e., the number of steps during which token $j$ resided in the cache, and define the final importance:
\[
i_j^{(\ell)} = \frac{\hat{c}_j^{(\ell)}}{e_j^{(\ell)}}\,.
\]

We aggregate attention over both heads and queries to obtain stable importance estimates. To safeguard essential references, we assign an importance of $+\infty$ to all tokens from the first frame—since it defines the global coordinate system—and similarly to camera and register tokens, as they play a critical role in ensuring accurate reconstruction. Our method overview is depicted in Fig~\ref{fig:method_overview}.
\subsection{Discussion}

In our approach, we introduce a \emph{per-layer} token budget $B$ that is applied directly to the KV cache. This budget can be flexibly set at inference time, requiring no architectural modifications or retraining. As a result, each layer's cache is limited to approximately $L_g \cdot B \cdot (d+d)$ (for keys and values), and this bound is independent of the total stream length $T$. Consequently, the computational cost of attention per step is determined by $B$, rather than by the cumulative history.

While Span3R \cite{wang20243d} also allows for dynamic adjustment of its long-term memory size at inference, our contribution lies in the introduction of per-layer budgeting within the transformer's internal KV cache, rather than simply enforcing a global memory constraint. This design enables more granular, layerwise control: as observed in the previous section, certain layers exhibit dense attention patterns while others are more selective. Furthermore, unlike Span3R’s external spatial memory—which requires re-projecting memory tokens into keys and values at every step—our per-layer KV cache design only computes queries for the current frame. This leads to faster per-step computation under the same token budget.

\section{Experiments}

In this section, we compare our proposed method against state-of-the-art approaches across 3D reconstruction, video depth estimation, and camera pose estimation benchmarks. We additionally conduct ablation studies to analyze the effectiveness of our eviction policy.  

\subsection{Implementation Details}
Our method is implemented by extending the StreamVGGT\cite{Zhuo2025Streaming4V} framework. We directly reuse the pretrained weights of StreamVGGT and apply our eviction mechanism during inference without retraining. The underlying architecture remains identical to StreamVGGT; the only modification is the insertion of token eviction in the global/temporal attention cache. In all experiments temperature for eviction is equal to 1.5 unless stated otherwise. Eviction applies only to \emph{global} self-attention KV caches; local/frame-wise attention is untouched. We disable FlashAttention in our experiments since it does not expose per-head attention maps, which are required for our analysis and visualizations, also in our experiments, StreamVGGT$^{\ast}$ donates StreamVGGT without FlashAttention. All experiments are performed on a single NVIDIA RTX6000-Ada GPU.

\subsection{Video Depth Estimation}
We follow the StreamVGGT evaluation protocol for video depth estimation. The task measures both the per-frame depth quality and the temporal consistency of predictions. Predicted depths are aligned with ground truth via per-sequence scale normalization. Results on Sintel and KITTI are summarized in Table~\ref{table:video_depth}, where we compare our method against StreamVGGT and other recent baselines. Our approach preserves accuracy at moderate budgets while offering tunable efficiency.

\begin{table}[t]
\caption{\textbf{Video Depth Evaluation (Sintel \& KITTI).} Metrics: Abs Rel $\downarrow$ (absolute relative error, lower is better) and $\delta < 1.25 \uparrow$ (percentage of predictions within a 1.25 factor of ground truth depth, higher is better).
First three methods take pairwise inputs, VGGT is Dense-view, and the others are streaming. “GA” marks methods with global alignment.}
\label{tab:viddepth_1col}
\centering
\setlength{\tabcolsep}{5pt}
\renewcommand{\arraystretch}{1.15}
\begin{tabular}{lcccc}
\hline
& \multicolumn{2}{c}{\textbf{Sintel}} & \multicolumn{2}{c}{\textbf{KITTI}} \\
\cline{2-5}
\textbf{Method} & Abs Rel$\downarrow$ & $\delta<1.25\uparrow$ & Abs Rel$\downarrow$ & $\delta<1.25\uparrow$ \\
\hline
DUSt3R\text{-}GA   & 0.656 & 45.2 & 0.144 & 81.3 \\
MASt3R\text{-}GA   & 0.641 & 43.9 & 0.183 & 74.5 \\
MonST3R\text{-}GA  & 0.378 & 55.8 & 0.168 & 74.4 \\
\textbf{VGGT}      & \textbf{0.298} & \textbf{68.1} & \textbf{0.061} & \textbf{97.0} \\
\hline
Spann3R            & 0.622 & 42.6 & 0.198 & 73.7 \\
CUT3R              & 0.421 & 47.9 & \textbf{0.118} & \textbf{88.1} \\
Point3R            & 0.452 & 48.9 & 0.136 & 84.2 \\
\textbf{StreamVGGT}& \textbf{0.323} & \textbf{65.7} & 0.173 & 72.1 \\
StreamVGGT$^{\ast}$& 0.325 & 65.4 & 0.174 & 72.1 \\
\hline

Ours ($B{=}0.1$) & 0.333 & 63.7 & 0.202 & 65.9 \\
Ours ($B{=}0.2$) & 0.335 & 63.5 & 0.209 & 62.0 \\
Ours ($B{=}0.3$) & 0.339 & 62.8 & 0.202 & 63.2 \\
Ours ($B{=}0.4$) & 0.337 & 62.4 & 0.193 & 65.9 \\
Ours ($B{=}0.6$) & 0.333 & 63.7 & 0.179 & 70.0 \\
Ours ($B{=}0.8$) & 0.330 & 65.0 & 0.174 & 71.9 \\
\hline
\end{tabular}
\label{table:video_depth}
\end{table}
\subsection{3D Reconstruction}

\begin{table*}[t]
\caption{\textbf{Quantitative 3D reconstruction results on 7-Scenes and NRGBD datasets.} Number of frames matches the
StreamVGGT experiment. Metrics: Accuracy (Acc, $\downarrow$), Completeness (Comp, $\downarrow$), Normal Consistency (NC, $\uparrow$). 
Reported as mean / median.
}
\label{tab:recon_main}
\centering
\setlength{\tabcolsep}{4.2pt}
\renewcommand{\arraystretch}{1.12}
\begin{tabular}{llcccccc|cccccc}
\hline
& & \multicolumn{6}{c|}{\textbf{7 scenes}} & \multicolumn{6}{c}{\textbf{NRGBD}} \\
\cline{3-14}
\textbf{Method} & \textbf{Type} &
\multicolumn{2}{c}{Acc$\downarrow$} & \multicolumn{2}{c}{Comp$\downarrow$} & \multicolumn{2}{c|}{NC$\uparrow$} &
\multicolumn{2}{c}{Acc$\downarrow$} & \multicolumn{2}{c}{Comp$\downarrow$} & \multicolumn{2}{c}{NC$\uparrow$} \\
& &
Mean & Med. & Mean & Med. & Mean & Med. &
Mean & Med. & Mean & Med. & Mean & Med. \\
\hline
DUSt3R-GA \cite{wang2024dust3r} & Pair-wise
& 0.146 & 0.077 & 0.181 & 0.067 & 0.736 & 0.839
& 0.144 & 0.019 & 0.154 & \textbf{0.018} & 0.870 & \textbf{0.982} \\
MASt3R-GA \cite{leroy2024grounding} & Pair-wise
& 0.185 & 0.081 & 0.180 & 0.069 & 0.701 & 0.792
& 0.085 & 0.033 & \textbf{0.063} & 0.028 & 0.794 & 0.928 \\
MonST3R-GA \cite{zhang2024monst3r} & Pair-wise
& 0.248 & 0.185 & 0.266 & 0.167 & 0.672 & 0.759
& 0.272 & 0.114 & 0.287 & 0.110 & 0.758 & 0.843 \\
\textbf{VGGT} \cite{Wang2025VGGTVG} & Dense-view
& \textbf{0.088} & \textbf{0.039} & \textbf{0.091} & \textbf{0.039} & \textbf{0.787} & \textbf{0.890}
& \textbf{0.073} & \textbf{0.018} & 0.077 & 0.021 & \textbf{0.910} & 0.990 \\
\hline
Spann3R \cite{wang20243d} & Streaming
& 0.298 & 0.226 & 0.205 & 0.112 & 0.650 & 0.730
& 0.416 & 0.323 & 0.417 & 0.285 & 0.684 & 0.789 \\
CUT3R \cite{Wang_2025_CVPR} & Streaming
& \textbf{0.126} & \textbf{0.047} & 0.154 & \textbf{0.031} & 0.727 & 0.834
& 0.099 & \textbf{0.031} & 0.076 & \textbf{0.026} & 0.837 & 0.971 \\
\textbf{StreamVGGT} & Streaming
& 0.129 & 0.056 & \textbf{0.115} & 0.041 & \textbf{0.751} & \textbf{0.865}
& \textbf{0.084} & 0.044 & \textbf{0.074} & 0.041 & \textbf{0.861} & \textbf{0.986} \\
StreamVGGT$^{\ast}$ & Streaming
& 0.133 & 0.059 & 0.117 & 0.043 & 0.750 & 0.863
& 0.084 & 0.044 & 0.074 & 0.041 & 0.861 & 0.986 \\
\hline
Ours ($B{=}0.10$) & Streaming
& 0.297 & 0.177 & 0.249 & 0.133 & 0.657 & 0.745
& 0.228 & 0.088 & 0.206 & 0.067 & 0.788 & 0.947 \\
Ours ($B{=}0.20$) & Streaming
& 0.167 & 0.067 & 0.151 & 0.049 & 0.737 & 0.851
& 0.190 & 0.075 & 0.176 & 0.072 & 0.815 & 0.958 \\
Ours ($B{=}0.30$) & Streaming
& 0.162 & 0.061 & 0.144 & 0.045 & 0.745 & 0.860
& 0.127 & 0.047 & 0.120 & 0.043 & 0.858 & 0.985 \\
Ours ($B{=}0.40$) & Streaming
& 0.142 & 0.066 & 0.124 & 0.048 & 0.749 & 0.861
& 0.095 & 0.046 & 0.094 & 0.039 & 0.860 & 0.983 \\
Ours ($B{=}0.60$) & Streaming
& 0.135 & 0.062 & 0.119 & 0.045 & 0.752 & 0.865
& 0.086 & 0.045 & 0.083 & 0.037 & 0.865 & 0.987 \\
Ours ($B{=}0.80$) & Streaming
& 0.132 & 0.059 & 0.117 & 0.043 & 0.750 & 0.863
& 0.084 & 0.044 & 0.078 & 0.038 & 0.862 & 0.986 \\
\hline
\end{tabular}
\end{table*}

\begin{table*}[h]
\caption{\textbf{3D Reconstruction with Longer Sequences.} 
7Scenes uses 2$\times$ frames, NRGBD uses 2.5$\times$ frames. 
Mem is the peak allocated GPU memory during inference.}
\label{tab:recon_long_2x_2p5x}
\centering
\setlength{\tabcolsep}{4pt}
\renewcommand{\arraystretch}{1.12}
\begin{tabular}{l|ccc|ccc|c|ccc|ccc|c}
\hline
& \multicolumn{7}{c|}{\textbf{7Scenes (2$\times$ frames)}} & \multicolumn{7}{c}{\textbf{NRGBD (2.5$\times$ frames)}} \\
\cline{2-15}
\textbf{Method} & Acc & Comp & NC & Acc$_\text{med}$ & Comp$_\text{med}$ & NC$_\text{med}$ & Mem (GB) & Acc & Comp & NC & Acc$_\text{med}$ & Comp$_\text{med}$ & NC$_\text{med}$ & Mem (GB) \\
\hline
Ours ($B{=}0.10$) & 0.106 & 0.081 & 0.717 & 0.059 & 0.036 & 0.826 & 7.68 & 0.175 & 0.134 & 0.787 & 0.103 & 0.086 & 0.949 & 7.57 \\
Ours ($B{=}0.20$) & 0.063 & 0.055 & 0.746 & 0.039 & 0.029 & 0.860 & 7.93 & 0.098 & 0.082 & 0.854 & 0.058 & 0.032 & 0.988 & 7.75 \\
Ours ($B{=}0.30$) & 0.058 & 0.052 & 0.748 & 0.032 & 0.024 & 0.861 & 8.19 & 0.100 & 0.082 & 0.855 & 0.054 & 0.036 & 0.987 & 7.97 \\
Ours ($B{=}0.40$) & 0.057 & 0.052 & 0.749 & 0.031 & 0.025 & 0.862 & 8.45 & 0.105 & 0.087 & 0.851 & 0.062 & 0.036 & 0.986 & 8.21 \\
Ours ($B{=}0.60$) & 0.057 & 0.052 & 0.749 & 0.031 & 0.025 & 0.861 & 8.98 & 0.104 & 0.085 & 0.850 & 0.060 & 0.037 & 0.986 & 8.69 \\
Ours ($B{=}0.80$) & 0.056 & 0.052 & 0.750 & 0.031 & 0.025 & 0.861 & 9.51 & 0.102 & 0.085 & 0.854 & 0.057 & 0.037 & 0.987 & 9.16 \\
\hline
StreamVGGT$^{\ast}$ & 0.056 & 0.052 & 0.750 & 0.030 & 0.025 & 0.861 & 9.75 & 0.101 & 0.084 & 0.855 & 0.057 & 0.037 & 0.987 & 9.35 \\
\hline
\end{tabular}
\end{table*}

\begin{table*}[t]
\caption{\textbf{3D Reconstruction with Ultra-Long Sequences.} 
7Scenes uses 8$\times$ frames, NRGBD uses 10$\times$ frames. }
\label{tab:recon_ultra_8x_10x}
\centering
\setlength{\tabcolsep}{4pt}
\renewcommand{\arraystretch}{1.12}
\begin{tabular}{l|ccc|ccc|c|ccc|ccc|c}
\hline
& \multicolumn{7}{c|}{\textbf{7Scenes (8$\times$ frames)}} & \multicolumn{7}{c}{\textbf{NRGBD (10$\times$ frames)}} \\
\cline{2-15}
\textbf{Method} & Acc & Comp & NC & Acc$_\text{med}$ & Comp$_\text{med}$ & NC$_\text{med}$ & Mem (GB) & Acc & Comp & NC & Acc$_\text{med}$ & Comp$_\text{med}$ & NC$_\text{med}$ & Mem (GB) \\
\hline
Ours ($B{=}0.01$) & 0.327 & 0.248 & 0.561 & 0.246 & 0.134 & 0.592 & 8.45 & 0.443 & 0.340 & 0.591 & 0.360 & 0.235 & 0.641 & 8.24 \\
Ours ($B{=}0.10$) & 0.047 & 0.034 & 0.688 & 0.026 & 0.013 & 0.791 & 9.39 & 0.085 & 0.050 & 0.819 & 0.059 & 0.022 & 0.973 & 8.77 \\
Ours ($B{=}0.20$) & 0.047 & 0.035 & 0.687 & 0.026 & 0.013 & 0.789 & 10.44 & 0.092 & 0.057 & 0.818 & 0.064 & 0.024 & 0.971 & 9.50 \\
Ours ($B{=}0.30$) & 0.046 & 0.034 & 0.687 & 0.026 & 0.014 & 0.789 & 11.51 & 0.099 & 0.060 & 0.813 & 0.067 & 0.023 & 0.968 & 10.40 \\
Ours ($B{=}0.40$) & 0.045 & 0.034 & 0.687 & 0.025 & 0.013 & 0.789 & 12.57 & 0.099 & 0.061 & 0.811 & 0.065 & 0.023 & 0.968 & 11.30 \\
Ours ($B{=}0.50$) & 0.045 & 0.033 & 0.688 & 0.025 & 0.013 & 0.790 & 13.63 & 0.098 & 0.060 & 0.813 & 0.065 & 0.023 & 0.969 & 12.20 \\
\hline
StreamVGGT$^{\ast}$ & 0.044 & 0.031 & 0.688 & 0.024 & 0.012 & 0.791 & 18.63 & 0.088 & 0.058 & 0.816 & 0.055 & 0.024 & 0.969 & 16.41 \\
\hline
\end{tabular}
\end{table*}

\begin{table*}[t]
\caption{\textbf{Camera Pose Estimation Evaluation on Sintel and TUM-dynamics.}
Lower is better. “Optim.” indicates test-time optimization; “Onl.” indicates online/streaming inference. OOM indicates out of memory.}
\label{tab:cam_pose_sintel_tum}
\centering
\setlength{\tabcolsep}{6pt}
\renewcommand{\arraystretch}{1.12}
\begin{tabular}{lcc|ccc|ccc}
\toprule
& \multicolumn{2}{c|}{\bf Settings} &
\multicolumn{3}{c|}{\bf Sintel} &
\multicolumn{3}{c}{\bf TUM-dynamics} \\
\cmidrule(lr){2-3}\cmidrule(lr){4-6}\cmidrule(l){7-9}
\textbf{Method} & \textbf{Optim.} & \textbf{Onl.} &
\textbf{ATE} $\downarrow$ & \textbf{RPE trans} $\downarrow$ & \textbf{RPE rot} $\downarrow$ &
\textbf{ATE} $\downarrow$ & \textbf{RPE trans} $\downarrow$ & \textbf{RPE rot} $\downarrow$ \\
\midrule
MASt3R-GA  & $\checkmark$ &  &
0.185 & 0.060 & 1.496 &
\textbf{0.038} & 0.012 & \textbf{0.448} \\
MonST3R-GA & $\checkmark$ &  &
\textbf{0.111} & 0.044 & 0.869 &
0.098 & 0.019 & 0.935 \\
Spann3R    &  & $\checkmark$ &
0.329 & 0.110 & 4.471 &
0.056 & 0.021 & 0.591 \\
CUT3R      &  & $\checkmark$ &
0.213 & 0.066 & \underline{0.621} &
0.046 & 0.015 & 0.473 \\
Point3R &  & $\checkmark$ &
0.351 & 0.128 & 1.822 &
0.075 & 0.029 & 0.642 \\
\midrule
StreamVGGT$^{\ast}$ & & $\checkmark$ & 0.232 & 0.095 & 0.722 & OOM & OOM & OOM \\
Ours ($B{=}0.9$) & & $\checkmark$ & 0.235 & 0.095 & 0.723 & OOM & OOM & OOM \\
Ours ($B{=}0.7$) & & $\checkmark$ & 0.259 & 0.099 & 0.717 & OOM & OOM & OOM\\
Ours ($B{=}0.5$) & & $\checkmark$ & 0.282 & 0.101 & 0.723 & 0.028 & 0.011 & 0.317 \\
Ours ($B{=}0.4$) & & $\checkmark$ & 0.287& 0.101 & 0.776 & 0.028 & 0.010 & 0.314 \\
Ours ($B{=}0.3$) & & $\checkmark$ & 0.282 & 0.103 & 0.852 & 0.027 & 0.012 & 0.314 \\
Ours ($B{=}0.2$) & & $\checkmark$ & 0.261 & 0.106 & 0.921 & 0.026 & 0.011 & 0.314 \\
Ours ($B{=}0.1$) & & $\checkmark$ & 0.244 & 0.088 & 0.931 & 0.027 & 0.012 & 0.314 \\
Ours ($B{=}0.01$) & & $\checkmark$ & 0.626 & 0.197 & 2.858 & 0.063 & 0.031 & 0.854 \\
\bottomrule
\end{tabular}
\end{table*}

Following CUT3R~\cite{Wang_2025_CVPR}, we evaluate 3D reconstruction on 7Scenes~\cite{Glocker2013RealtimeRC} and NRGBD~\cite{sturm2012benchmark}, reporting accuracy (Acc), completeness (Comp), and normal consistency (NC). The standard setup uses sparse inputs (3–5 frames for 7-Scenes, 2–4 for NRGBD), with extended tests on longer sequences: 2$\times$/2.5$\times$ frames and ultra-long 8$\times$/10$\times$ frames. In addition to reconstruction quality, we also report peak GPU memory usage.  

As shown in Table~\ref{tab:recon_main}, our method matches StreamVGGT under strict budgets. With longer sequences (Table~\ref{tab:recon_long_2x_2p5x}), StreamVGGT memory rises sharply (up to 9.35 GB on NRGBD and 9.75 GB on 7Scenes), while our method maintains comparable accuracy at lower cost. For ultra-long sequences (Table~\ref{tab:recon_ultra_8x_10x}), StreamVGGT requires 16.41 GB (NRGBD) and 18.63 GB (7Scenes), but our method achieves similar accuracy with nearly half the memory. The best trade-off appears at $B=0.1$, which even outperforms StreamVGGT on NRGBD while using far less memory. Overall, token eviction enables scalable streaming reconstruction with strong accuracy–efficiency balance. Importantly, in comparing these 3 tables for 3D reconstruction, we can say that with an increasing number of frames sampled for reconstruction and a low budget (e.g., $B=0.1$), our method not only reduces memory usage but can also achieve superior performance compared to the baseline. While it is true that inference time increases with the number of frames, token eviction helps recover some of this overhead by bounding the effective cache size. We also report the per-frame latency in Table~\ref{tab:latency}.

\begin{figure*}[t]
  \centering
  \begin{subfigure}{0.24\textwidth}
    \includegraphics[width=\linewidth]{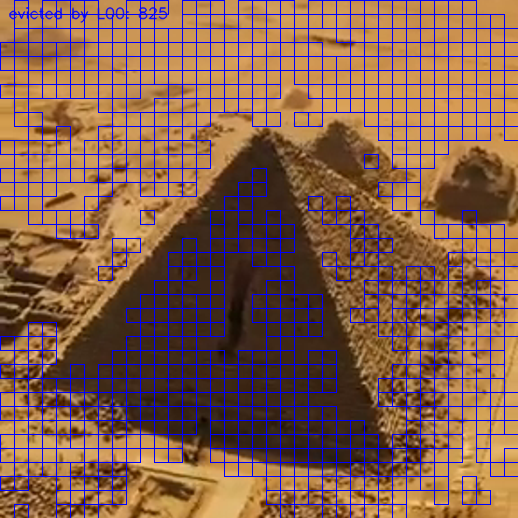}
    \caption{Frame 2}
  \end{subfigure}
  \begin{subfigure}{0.24\textwidth}
    \includegraphics[width=\linewidth]{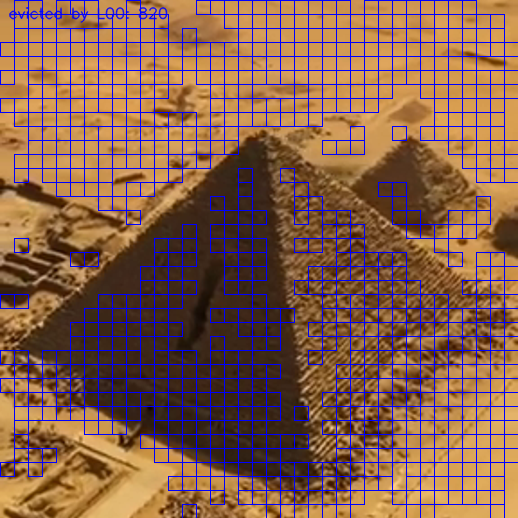}
    \caption{Frame 3}
  \end{subfigure}
  \begin{subfigure}{0.24\textwidth}
    \includegraphics[width=\linewidth]{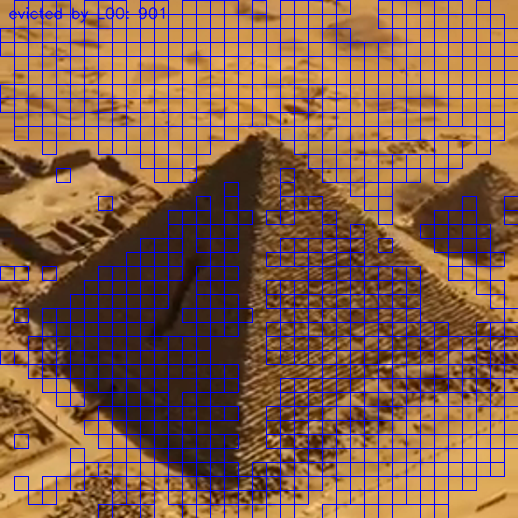}
    \caption{Frame 4}
  \end{subfigure}
   \begin{subfigure}{0.24\textwidth}
    \includegraphics[width=\linewidth]{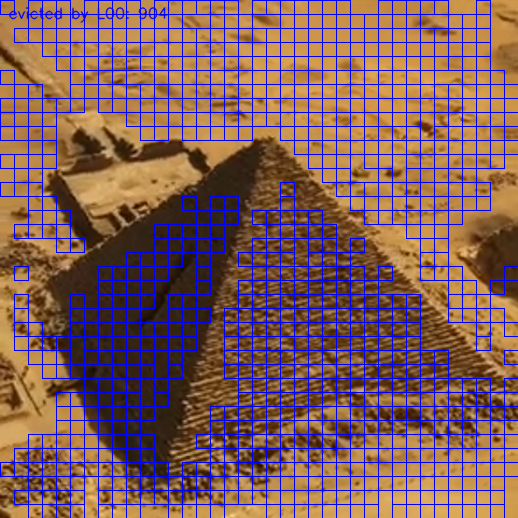}
    \caption{Frame 5}
  \end{subfigure}


  \caption{%
    \textbf{Token eviction masks for first layer across frames 2–5 of an 11-frame sequence.}
    Each subplot shows which tokens are retained or evicted at that frame. Blue-bordered squares indicate evicted tokens.
  }

  \label{fig:eviction_grid_layer_i}
\end{figure*}

\subsection{Camera Pose Estimation}
We evaluate camera pose estimation on Sintel~\cite{butler2012naturalistic} and TUM-dynamics~\cite{sturm2012benchmark} datasets, following the setup of MonST3R~\cite{zhang2024monst3r} and CUT3R~\cite{Wang_2025_CVPR}. Metrics include Absolute Translation Error (ATE), Relative Translation Error (RPE$_{trans}$), and Relative Rotation Error (RPE$_{rot}$), computed after Sim(3) Umeyama alignment~\cite{zhang2024monst3r}. Unlike several baselines that require test-time optimization, our method runs fully feed-forward without post-processing. Results in Table~\ref{tab:cam_pose_sintel_tum} show that our streaming model achieves competitive performance with online methods while retaining a performance gap relative to optimization-heavy approaches.


\subsection{Ablation Studies}
We conduct additional experiments to analyze the components of our eviction strategy:

\subsubsection{Latency under different sequence lengths} We evaluate per-frame latency on 7-Scenes under both the standard (1$\times$) and ultra-long (8$\times$) regimes, and find that token eviction helps stabilize inference time as sequence length increases (Table~\ref{tab:latency}).

\subsubsection{Eviction policies} 
We compare our scoring-based eviction with two alternatives: random eviction and uniform budget allocation across layers. On 7Scenes with 2$\times$ frames (Table~\ref{tab:ablation_7scenes_2f}), our method achieves the lowest errors across budgets in the accuracy metric, particularly at $B{=}0.1$ where random eviction degrades sharply in accuracy and completeness. These results confirm that our scoring method is effective and more robust under tight budgets, preserving reconstruction quality while maintaining bounded memory.

\subsubsection{Qualitative eviction masks} 
We visualize tokens evicted from the global transformer’s first-layer cache across different input frames of a sample sequence (Fig.~\ref{fig:eviction_grid_layer_i}). The results show that evicted tokens largely correspond to background or low-texture regions, while tokens covering salient geometry are preserved. This behavior is consistent with our design objective and supports the quantitative gains reported in earlier tables.

\begin{table}[!t]
\caption{Per-frame latency (s). on 7scenes across sequences.}
\label{tab:latency}
\centering
\begin{tabular}{lcc}
\hline
\textbf{Method / $B$} & \textbf{1$\times$} & \textbf{8$\times$} \\
\hline
StreamVGGT    & 0.107 & 0.142 \\
StreamVGGT$^{\ast}$     & 0.141 & 0.475 \\
Ours ($B{=}0.7$)      & 0.158 & 0.491 \\
Ours ($B{=}0.5$)      & 0.150 & 0.416 \\
Ours ($B{=}0.3$)      & 0.148 & 0.315 \\
Ours ($B{=}0.1$)      & 0.131 & 0.187 \\
\hline
\end{tabular}
\end{table}

\begin{table}[!t]
\centering
\caption{Ablation on 7Scenes with 2$\times$ frames. Uniform Budget temperature is 100.}
\resizebox{\columnwidth}{!}{%
\begin{tabular}{rcccccc}
\toprule
\bf Budget &
\multicolumn{2}{c}{\bf Ours} &
\multicolumn{2}{c}{\bf Uniform Budget} &
\multicolumn{2}{c}{\bf Random Evict} \\
\cmidrule(lr){2-3}\cmidrule(lr){4-5}\cmidrule(l){6-7}
& Acc$\downarrow$ & Comp$\downarrow$ & Acc$\downarrow$ & Comp$\downarrow$ & Acc$\downarrow$ & Comp$\downarrow$ \\
\midrule
0.50 & 0.0570 & 0.0520 & 0.0570 & 0.0520 & 0.058 & 0.053 \\
0.30 & 0.0580 & 0.0520 & 0.0580 & 0.0520 & 0.067 & 0.056 \\
0.10 & 0.1060 & 0.0810 & 0.1440 & 0.0770 & 0.239 & 0.230 \\
\bottomrule
\end{tabular}
}
\label{tab:ablation_7scenes_2f}
\end{table}

\section{Conclusion}
We addressed the memory scalability challenge of StreamVGGT by introducing a simple yet effective token eviction strategy. Our approach preserves critical tokens while discarding redundant ones, allowing reconstruction and depth estimation to scale efficiently to long sequences. Extensive experiments on 7Scenes, NRGBD, Sintel, KITTI, and TUM-dynamics show that our method maintains accuracy under strict memory budgets and achieves superior trade-offs between efficiency and performance, particularly in ultra-long sequences. Ablation studies further validate the importance of our eviction design and its impact on stability and latency. We believe this work provides a practical step toward scalable and real-time streaming 3D perception models.

\addtolength{\textheight}{-2cm}   



\bibliographystyle{IEEEtran}   
\bibliography{ref} 

\end{document}